\documentclass{article}
\usepackage[final]{neurips_2022}

\usepackage{natbib}
\usepackage[utf8]{inputenc} 
\usepackage[T1]{fontenc}    
\usepackage{url}            
\usepackage{booktabs}       
\usepackage{amsfonts}       
\usepackage{nicefrac}       
\usepackage{graphicx}
\graphicspath{{ArXiv/figures/}}
\usepackage{fullpage}
\usepackage{enumitem}
\usepackage{subcaption}
\usepackage{amssymb,amsmath,amsthm}
\usepackage[dvipsnames]{xcolor}
\definecolor{DarkGreen}{rgb}{0.1,0.5,0.1}
\usepackage[colorlinks = true,
            linkcolor = DarkGreen,
            urlcolor  = DarkGreen,
            citecolor = DarkGreen,
            anchorcolor = DarkGreen]{hyperref}
\usepackage{booktabs,array,dcolumn}
\usepackage{siunitx}
\usepackage{xcolor}
\definecolor{myblue}{rgb}{0, .5, 1}
\def\shownotes{1}  
\ifnum\shownotes=1
\newcommand{\authnote}[2]{{$\ll$\textsf{\footnotesize #1: #2}$\gg$}}
\else
\newcommand{\authnote}[2]{}
\fi

\newcolumntype{d}{D{.}{.}{2.3}}
\newcolumntype{C}{>{\centering}p}

\theoremstyle{definition}

\title{Improving Molecule Properties Through 2-Stage VAE}

\author{Chenghui Zhou}
\author{Barnabás Póczos}

\author{%
  Chenghui Zhou\\
  Department of Machine Learning\\
  Carnegie Mellon University\\
  \texttt{chenghuz@andrew.cmu.edu} \\
  \And
   Barnabás Póczos\\
   Department of Machine Learning\\
  Carnegie Mellon University\\
  \texttt{bapoczos@cs.cmu.edu} \\
}

\begin{document}

\maketitle

\begin{abstract}
    Variational autoencoder (VAE) is a popular method for drug discovery and there had been a great deal of architectures and pipelines proposed to improve its performance. But the VAE model itself suffers from deficiencies such as poor manifold recovery when data lie on low-dimensional manifold embedded in higher dimensional ambient space and they manifest themselves in each applications differently. The consequences of it in drug discovery is somewhat under-explored. In this paper, we study how to improve the similarity of the data generated via VAE and the training dataset by improving manifold recovery via a 2-stage VAE where the second stage VAE is trained on the latent space of the first one. We experimentally evaluated our approach using the ChEMBL dataset as well as a polymer datasets. In both dataset, the 2-stage VAE method is able to improve the property statistics significantly from a pre-existing method.
\end{abstract}
\vspace{-12pt}
\section{Introduction}
\vspace{-8pt}
The use of generative models in the domain of drug discovery has recently seen rapid progress. These methods can leverage large-scale molecule archives describing the structure of existing drugs to synthesize novel molecules with similar properties as potential candidates for future drugs \citep{duvenaud2015convolutional, liu2018constrained, segler2018generating, you2018graph, jin2018junction, jin2020hierarchical, polykovskiy2020molecular, jin2020multi, satorras2021n}. There are two common ways of representing the structure of molecules SMILES strings \citep{weininger1988smiles} and molecular graphs \citep{bonchev1991chemical}. Graph neural networks can make effective use of the rich molecular graph representations by taking into account the atoms, edges and other structural information. SMILES strings convey less information about the molecular structure, but are more compatible with sequence models such as RNNs. Being able to generate valid molecules is the first step to machine learning drug discovery and various solutions have been proposed. For example, GNN methods \citep{liu2018constrained, jin2020hierarchical, simonovsky2018graphvae} can constraint the output space based on the chemical rules and SMILES-based \citep{gomez2018automatic, blaschke2018application} approaches benefit from the abundant molecular data. 

Despite the valid molecule outputs, the properties of the generated molecules, such as drug-likeness(QED) \citep{bickerton2012quantifying}, Synthetic Accessibility Score (SA) \citep{ertl2009estimation} and molecular weight (MW) etc., are critical factors that decide whether they can be  synthesized in a laboratory and be effective in real world applications. In order to learn to generate molecules that fulfill these properties, researchers curated molecule datasets for targeted purposes, such as ChEMBL \citep{mendez2019chembl} and ZINC \citep{irwin2005zinc}. The motivation is that by learning from a curated set of molecules, the generative models will learn to generate similar ones. Benchmark metrics \citep{polykovskiy2020molecular} are created to measure how similar the generated molecules are to the target dataset but the results show that we still have more room to improve on this front. Learning to generate molecules that exhibit {\it similar} molecular properties to those in the target dataset is a prerequisite to achieve the desired properties in the generated molecules. 

In this paper, we introduce an easy-to-implement step to the VAE approach -- training an additional VAE to generate the latent for the first-stage VAE-- to improve the property metrics of the generated molecules by mitigating the manifold recovery problem. Experimentally, we first show how this approach can enhance the manifold recovery for synthetic data. We then evaluate our method in two domains using the ChEMBL dataset and the polymer datasets \citep{st2019message}. In both settings, the 2-Stage VAE approach is able to enhance the property statistics of the generated molecule set by bringing them closer in distribution to the training set. 

\vspace{-8pt}
\section{Related Work}
\vspace{-8pt}
We structure our discussing based on the type of molecular representation underlying the individual methods. Most current approaches fall into one of the following families -- namely, the SMILES strings approach, the molecular graph approach and the 3D point set approach. Many approaches have been proposed to generate molecules as SMILES strings \citep{segler2018generating, gomez2018automatic}. \cite{kusner2017grammar, dai2018syntax} took advantage of the syntax of the SMILES strings to constrain the output of the VAE model in order to improve validity of the generated molecules. Other approaches to generate SMILES strings include generative adversarial model \citep{kadurin2017drugan, prykhodko2019novo, guimaraes2017objective}. Molecular graphs carry more information about the molecular structures than the SMILES string format and GNN can effectively incorporate the additional information into the learning process \citep{duvenaud2015convolutional, liu2018constrained}. \cite{jin2018junction} proposed to generate molecular graph in two steps -- generate the tree-structured scaffolds first, and then combine with the substructures to form molecules. \cite{jin2020hierarchical} improved upon this prior result and proposed to generate via substructures in a course-to-fine manner to adapt to bigger molecules, such as polymers. \cite{satorras2021n} introduced an equivariant graph neural network to apply on molecular graphs.  3D representations of molecules are gaining traction in the research communities as they provide additional spatial information of the molecules \citep{gebauer2019symmetry, gebauer2022inverse, luo20213d, hoogeboom2022equivariant}. However, none of the methods use the VAE framework. Our paper is limited to improving the VAE approaches. Other generative approaches to drug discovery include generative adversarial model \citep{kadurin2017drugan, prykhodko2019novo, guimaraes2017objective} and diffusion models \citep{hoogeboom2022equivariant, xu2022geodiff}.




\vspace{-8pt}
\section{Method}
\vspace{-8pt}
\begin{figure}
\centering
\includegraphics[width=0.9\textwidth]{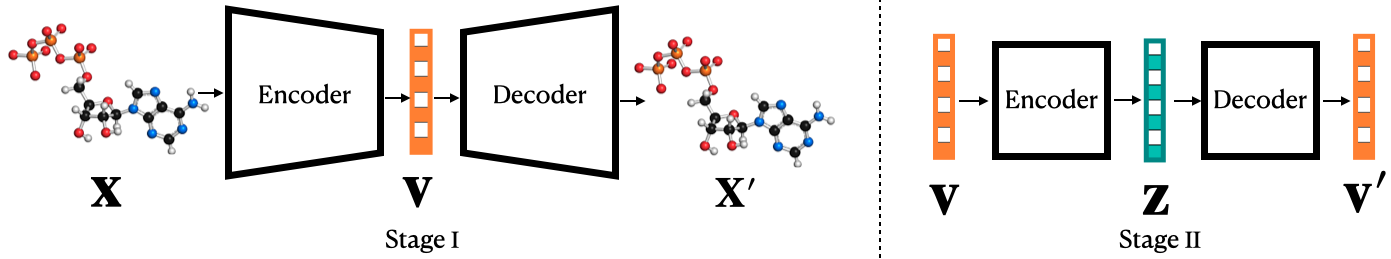}  
\caption{Overview of 2-stage VAE. In the first stage, the VAE trains with the molecule data $x_i$ and obtain the latent variables $v_i$'s from each of the training points. $v_i$'s become the input of the second stage VAE. The second stage VAE's input dimension is equal to the output dimension. During sampling time, we sample $z ~ \mathcal{N}(0, I)$ and obtain $v$ through the second stage decoder. It is then used as the latent variable for the first stage VAE to be decoded into a molecule.}
\label{fig:2_stage_demo}
\vspace{-15pt}
\end{figure}

The VAE famework \citep{kingma2013auto} has enabled great success in the image generation domain and more recently VAE based approaches have become a popular approach for addressing the molecule generation problem. Many sophisticated architectures have been proposed to adapt the VAE approach to molecular data \citep{kusner2017grammar, dai2018syntax, jin2019hierarchical, satorras2021n}. However, adapting the underlying neural architecture does not remedy VAE's learning deficiency in manifold recovery \citep{dai2019diagnosing, koehler2021variational}. In the case of high-dimensional data that lies on low-dimensional manifold such as images and molecular representations, \cite{koehler2021variational} found that the VAE is not guaranteed to recover the manifold where the nonlinear data lie. The 2-stage method can improve manifold recovery as demonstrated in a synthetic experiment (Figure \ref{fig:synthetic_sphere_mol}) and further enhance the performance of a pre-existing model. 
\vspace{-5pt}
\subsection{Variational Autoencoder}
\vspace{-5pt}
The variational inference framework assumes that the data $x$ is generated from a latent variable $z \sim p(z)$. The prior $p(z)$ is assumed to be a multivariate standard normal distribution in the application of a VAE. A VAE model seeks to maximize the likelihood of the data, denoted as $\log p_\theta(x) = \log \int p(z) p_{\theta}(x|z) d z$ where $\theta$ denotes the generative parameters. However, the marginalization is intractable in practice due to the inherent complexity of the generator, or the decoder,  thus an approximation of the objective is needed. 
Let $\phi$ be the variational parameters, the VAE model consists of a tractable encoder $q_{\phi}(z|x)$ and a decoder $p_{\theta}(x|z)$. Together, they approximate a lower bound to the log likelihood of the data. Ideally, by optimizing this lower bound we aim to increase the likelihood. This approximation enables the efficient posterior inference of the latent variable $z$ given the output $x_i$ and for marginal inference of the output variable $x$. The objective function of VAE is:
\vspace{-2pt}
\begin{equation}
\mathcal{L}(\theta, \phi; x) = -D_{KL}(q_{\phi}(z|x) \,||\, p(z)) + \mathbb{E}_{q_{\phi}(z|x)}[\log p_{\theta}(x | z)] \leq \log p_\theta(x)\end{equation}
\vspace{-2pt}
For generation, latent variable $z_i$ is sampled from the prior $p(z)$ which is a multivariate standard normal and the decoder transforms $z_i$ into the output $x_i$. 
\vspace{-5pt}
\subsection{2-Stage VAE}
\vspace{-5pt}
Despite its widespread use, VAE in its original form has many known flaws. Particularly, in the case where the data lies on low-dimensional manifold embedded in a high-dimensional ambient space. \cite{dai2019diagnosing} hypothesized that
training a VAE with a fixed decoder variance could add additional noise to the output, while training with tunable decoder variance, the decoder variance has a tendency approach zero and the model will learn the correct manifold but not the correct density. In practice this can lead to low-quality output images in comparison to models such as GANs. The implications of the finding extends to molecular data as well. Subsequently, \cite{dai2019diagnosing} proposed a 2-stage VAE approach to enhance the manifold and density recovery of existing VAE approaches. The reasoning was that the first stage VAE with tunable decoder variance learns the low-dimensional manifold the data lies on as the decoder variance goes to zero and the probability mass collapses onto the correct low-dimensional manifold. The second stage VAE is constructed with its latent dimension equal to the output dimension and is theorized to recover the density when the ambient dimension is equal to the intrinsic dimension. Their algorithm visibly improved the appearance of the generated images but the claim of manifold or density recovery cannot be easily verified with image data. 

\cite{koehler2021variational} showed empirically and analytically that when the data is a nonlinear function of the latent variables, neither the manifold nor the density is guaranteed to be recovered by the first stage VAE. This could provide an explanation as to why the generated molecules from the VAE are dissimilar in properties to the training datset. Even though this conclusion rendered the reasoning behind a 2-stage VAE invalid, the algorithm itself is not without merits. As we will demonstrate in the following synthetic experiment, a 2-stage VAE can improve manifold recovery. This could significantly improve the properties of the generated molecules as we will demonstrate in the experiments.

\begin{figure}
\centering
\begin{subfigure}{0.33\linewidth}
  \centering
  \includegraphics[width=\textwidth]{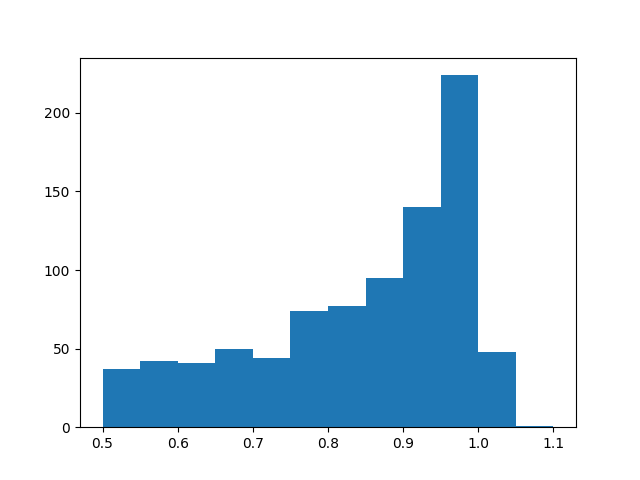}
  \caption{Stage 1}
  \label{fig:sub1}
\end{subfigure}%
\begin{subfigure}{0.33\linewidth}
  \centering
  \includegraphics[width=\textwidth]{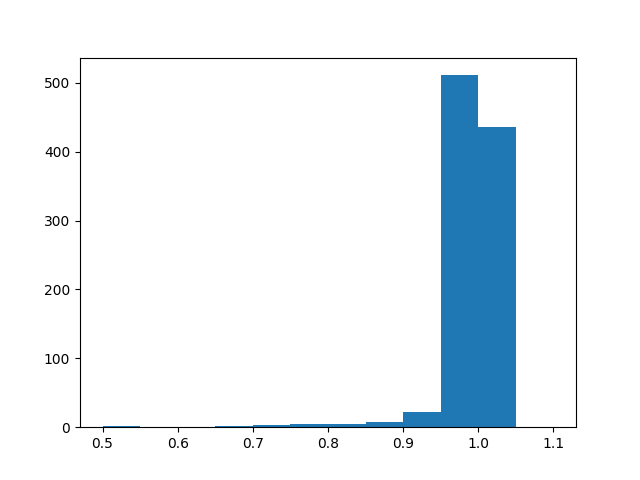}
  \caption{Stage 2}
  \label{fig:sub2}
\end{subfigure}
\begin{subfigure}{0.33\linewidth}
  \centering
  \includegraphics[width=\textwidth]{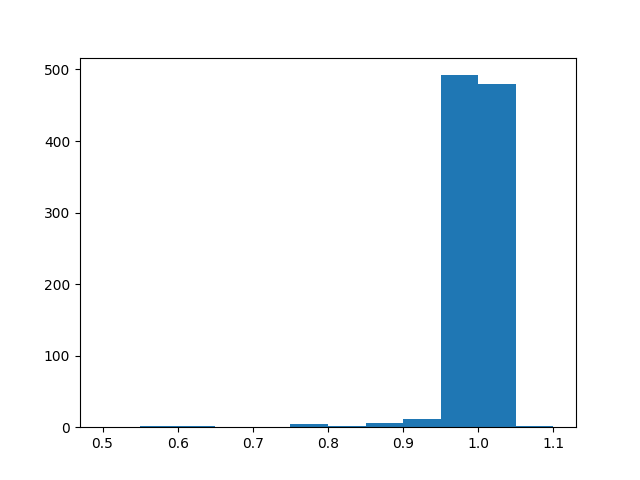}
  \caption{Stage 3}
  \label{fig:sub3}
\end{subfigure}
\caption{Multi-stage VAE on synthetic data. The $x$-axis represents the norm of the data point and the $y$-axis represents the number of data points that are of $x$ distance away from the unit sphere center. The figures across different stages of VAE training shows that the sphere surface is eventually recovered and improved starting from stage 2.}
\label{fig:synthetic_sphere_mol}
\vspace{-10pt}
\end{figure}

\paragraph{Synthetic Experiment} We show that a 2-stage VAE setup improves the recovery of the manifold. We demonstrate this in a synthetic experiment with data generated from a ground-truth manifold (see Figure \ref{fig:synthetic_sphere_mol}). We generate data from a 2-dimensional unit sphere (the norms of all the generated data points are 1). The 3-dimensional vectors are then padded with 16 dimensions of zeros to embed the data in a higher ambient space. In this case, the intrinsic dimension of the data is 2 and the ambient dimension is 17. We trained on this data in 3 stages -- meaning, the latents from the previous stage are used for training in the next stage. For the second and the third stage VAE, the latent dimension is set to be the same as their input (Figure \ref{fig:2_stage_demo}). The decoder variance is tunable for all stages and the decoder variance of the first stage approaches 0 upon convergence. During sampling, the output of the later stage VAE's becomes the latent of the previous stage VAE. The last-stage VAE's latents are sampled from standard normal distribution. We sample 1000 data points to visualize the results in the histograms. They show that the VAE in the first stage {\it does not} recover the manifold and many of the generated data points fall {\it inside} of the sphere, echoing the finding by \cite{koehler2021variational}. In the second and third stage, we see that more data points fall {\it on} the sphere, indicating the recovery of the manifold. 

\paragraph{Application on Molecule Generation} How improving manifold recovery of the generative model would benefit molecule generations does not have a straightforward answer. Evaluation metrics on molecule generation are multifaceted. Validity alone provides only a shallow examination on the quality of the generated molecules and other property metrics need to be considered to evaluate the generated molecules for real world applications. We provide empirical studies in the experiment section to study the suitability of a 2-stage VAE approach in the molecule generation domain by providing evaluation results on sample quality, structural as well as property statistics \citep{polykovskiy2020molecular}. We found that the 2-stage VAE helps to generate molecules that are more similar in properties to the ones in the training dataset. We present the precise steps to train a 2-stage VAE \citep{dai2019diagnosing}:

\begin{enumerate}
    \item Train a VAE on the molecular dataset $\{x_i | i = 1, 2, \dots n \}$ with architectures of your choice. Upon convergence, save the latent vectors $v_i \sim q_{\phi} (v | x_i) $, for all the molecules in the dataset. The tunable decoder covariance requirements are satisfied with decoders that follow multinomial distribution such as in molecule generations, the decoder variance approaching zero is equivalent to the probability mass concentrating on one choice;
    \item With $v_i$'s as input, train the second stage VAE with tunable decoder variance. We denote the latent of the second stage VAE as $z$. In this paper, we use feed-forward architectures for both the decoder $p_{\theta^{'}} (v|z)$ and the encoder $q_{\phi{'}} (z|v)$. The dimension of $v$ is equal to the dimension $z$ for maximum power. 
    \item During the generation process, sample the latent of the last stage VAE by $z \sim \mathcal{N}(0, I)$. Obtain the output from the second stage decoder $v_i \sim p_{\theta^{'}}(v | z)$ as the latent for the first stage VAE. And get the molecule sample $x$ from the first stage decoder via $x \sim p_{\theta}(x|v)$.
\end{enumerate}

One way to interpret this method is that while the first-stage VAE learns a mapping between the latent representations and the molecular data, the second-stage VAE learns to generate latent variables from the distribution of the latent representations of the dataset.


\vspace{-5pt}
\section{Experiments}
\vspace{-8pt}
In this section, we explore in detail how the 2-stage VAE improves the generated molecules. We adopt two model architectures -- hierarchical GNN and character-level RNN -- to compare the outcomes of a 2-stage and 3-stage VAE on different model architectures. We adopt a GAN-based model as an additional comparison. We conducted experiments on two molecule datasets -- ChEMBL \citep{mendez2019chembl} and polymers \citep{st2019message} for a comprehensive study of the method. 


We introduce the two datasets used in the experiments -- ChEMBL dataset \cite{mendez2019chembl} and polymers dataset \cite{st2019message}. Details on our benchmark metrics are in Appendix \ref{benchmark metric}

\begin{itemize}
\item  \textbf{ChEMBL Dataset}\citep{mendez2019chembl} consists of 1,799,433 bioactive molecules with drug-like properties. It is split into training, testing, validation, test scaffold and validation scaffold dataset containing 1,463,775, 81,321, 8,321, 86,508 and 86,507 molecules respectively.
    
\item \textbf{The Polymer Dataset}\citep{st2019message} contains 86,353 polymers and it's divided into training, test and validation set that contains 76,353, 5000 and 5000 molecules each. There is no scaffold split for the polymers dataset. Polymers generally have heavier weight than the molecules in the ChEMBL dataset and the dataset size is smaller.
\end{itemize}

We use hierearchical GNN (HGNN) \citep{jin2019hierarchical} and vanilla RNN (RNN) \citep{polykovskiy2020molecular} as the first stage VAE and a GAN-based model, latent GAN \citep{prykhodko2019novo}, as baseline:

  
\begin{table}[h!]
\centering
  \resizebox{\textwidth}{!}{  
  \begin{tabular}{c|cccc|ccc|cccc}
 &\multicolumn{4}{c}{Sample Quality } & \multicolumn{3}{c}{Structural Statistics } & \multicolumn{4}{c}{Property Statistics } \\
  Stage \# & Valid $\uparrow$ & Unique $\uparrow$ & Novelty $\uparrow$ & FCD $\downarrow$ & SNN $\uparrow$ & Frag $\uparrow$ & Scaf $\uparrow$ & LogP $\downarrow$ & SA $\downarrow$ & QED $\downarrow$ & MW $\downarrow$\\
  \hline
  \hline
  HGNN\#1 & $\textbf{1.0}$ & $\textbf{1.0}$ & $0.99$ & $5.1$ & $\textbf{0.42}$ & $0.97$ & $\textbf{0.46}$ & $0.92{\tiny 0.016}$ & $0.070{\tiny 4.3\mathrm{e}{-3}}$ & $0.024{\tiny 9.5\mathrm{e}{-4}}$ & $68.8{\tiny 0.83}$\\
  HGNN\#2 &$\textbf{1.0}$ &$\textbf{1.0}$ &$0.99$ &$\textbf{1.1}$ &$0.41$ &$\textbf{1.0}$ &$0.43$ & $0.095{\tiny 0.019}$ & $\textbf{0.069}{\tiny 5.8\mathrm{e}{-3}}$ & $\textbf{0.0067}{\tiny 1.0\mathrm{e}{-3}}$ & $5.0{\tiny 0.72}$\\
  HGNN\#3 & $\textbf{1.0}$ &$\textbf{1.0}$ &$\textbf{1.0}$ &$1.2$ &$0.41$ &$\textbf{1.0}$ & $\textbf{0.46}$ & $\textbf{0.059}{\tiny 4.5\mathrm{e}{-3}}$ & $0.069{\tiny 6.3\mathrm{e}{-3}}$ & $0.016{\tiny 1.6\mathrm{e}{-3}}$ & $7.7{\tiny 0.42}$ \\
  \hline
  RNN\#1 & $0.86$ &$\textbf{1.0}$ &$\textbf{1.0}$ &$1.84$ &$0.38$ &$\textbf{1.0}$ &$0.38$ & $0.088{\tiny 7.8\mathrm{e}{-3}}$ &$0.25{ \tiny 7.8\mathrm{e}{-3}}$ &$0.0088{\tiny 1.6\mathrm{e}{-3}}$ &$3.2{\tiny 0.55}$  \\
  RNN\#2 & $0.87$ &$\textbf{1.0}$ &$\textbf{1.0}$ &$1.86$ &$0.38$ &$\textbf{1.0}$ &$0.36$ &$0.099{\tiny 5.5\mathrm{e}{-3}}$ &$0.27{\tiny 7.7\mathrm{e}{-3}}$ & $0.0099{\tiny 1.5 \mathrm{e}{-3}}$ &$\textbf{2.8}{\tiny 0.29}$ \\
  \hline
  LatentGan &$0.77$ &$0.98$ &$0.99$ & $17.3$ & $0.34$ &$0.68$ &$0.21$ & $0.69{\tiny 0.019}$ & $0.63{\tiny 7.3\mathrm{e}{-3}}$ &$0.047{\tiny 2.0\mathrm{e}{-3}}$ &$27.2{\tiny 0.88}$
\end{tabular}
}
  \caption{Properties of the generated molecules trained on the ChEMBL dataset.}
  \vspace{-10pt}
  \label{tab:chembl property table}
\end{table}

\begin{table}[h!]
  \begin{center}
  \resizebox{\textwidth}{!}{  
  \begin{tabular}{c|cccc|ccc|cccc}
  &\multicolumn{4}{c}{Sample Quality } & \multicolumn{3}{c}{Structural Statistics } & \multicolumn{4}{c}{Property Statistics }\\
  Stage \# & Valid $\uparrow$ & Unique $\uparrow$ & Novelty $\uparrow$ & FCD $\downarrow$ & SNN $\uparrow$ & Frag $\uparrow$ & Scaf $\uparrow$ & LogP $\downarrow$ & SA $\downarrow$ & QED $\downarrow$ & MW $\downarrow$\\
  \hline
  \hline
  HGNN\#1 & \textbf{1.0} & \textbf{1.0} & 0.57 &0.62 & 0.67 &0.98 &0.37 &1.3{\tiny $0.030$} &0.089{\tiny $3.0\mathrm{e}{-3}$} &0.020{\tiny $1.2\mathrm{e}{-3}$} & 72.2{\tiny $1.42$} \\
  HGNN\#2 & \textbf{1.0} & \textbf{1.0} & 0.51 & \textbf{0.27} & \textbf{0.69} & \textbf{0.99} & 0.37& \textbf{0.10}{\tiny $0.033$} &0.031{\tiny $3.3\mathrm{e}{-3}$} &0.0041{\tiny $9.5\mathrm{e}{-4}$} &\textbf{7.7}{\tiny $1.1$} \\
  HGNN\#3 & \textbf{1.0} &\textbf{1.0} &0.52 &0.29 &\textbf{0.69} &\textbf{0.99} &0.38 &0.24{\tiny $0.017$} &\textbf{0.024}{\tiny $4.1\mathrm{e}{-3}$} &\textbf{0.0024}{\tiny $2.9\mathrm{e}{-4}$} &9.4{\tiny $2.3$} \\
  \hline
  RNN\#1 & 0.53 & 0.99 & 0.13 & 1.6 & 0.69 & 0.87 & 0.50 & 2.6{\tiny $0.011$} &0.31{\tiny $9.2\mathrm{e}{-3}$} & 0.047{\tiny $9.4\mathrm{e}{-4}$} &178.4{\tiny $1.2$}  \\
  RNN\#2 &0.53 & \textbf{1.0} & 0.13 &1.5 & 0.69 &0.87 &\textbf{0.51}& 2.5{\tiny $0.011$} & 0.31{\tiny $3.2\mathrm{e}{-3}$} & 0.044{\tiny $6.9\mathrm{e}{-4}$} & 176.1{\tiny $0.68$}  \\
  \hline
  LatentGan & 0.94 & \textbf{1.0} & \textbf{0.82} & 0.51 & 0.66 & \textbf{0.99} & 0.33 & 0.26{\tiny $0.031$} &0.041{\tiny $3.3\mathrm{e}{-3}$} & 0.0029{\tiny $4.37\mathrm{e}{-4}$} &11.8{\tiny $1.7$} 
\end{tabular}
}
\end{center}
  \caption{Properties of the generated molecules trained on the polymers dataset.}
  \vspace{-15pt}
  \label{tab:polymer property table}
\end{table}

We sampled 10,000 molecules from each model to generate the results in Table \ref{tab:chembl property table} and Table \ref{tab:polymer property table}. We included sample quality, structural and property statistics. The numbers in the tables are averaged over 6 sets of samples generated with 6 different random seeds from the model. We included the standard deviations for the property statistics but eliminated the rest as those are below 0.01. 

On both datasets, the HGNN\#2 improves upon the first stage by many folds on property statistics. The most notable improvement from the ChEMBL dataset is the QED (from 0.024 to 0.0067), MW (from 68.8 to 5.0) and LogP (0.92 to 0.059). On the polymer dataset, the second stage VAE improves significantly across all metrics -- from 72.2 to 7.7 on MW, 0.020 to 0.0024 on QED, 0.089 to 0.031 on SA and 1.3 to 0.1 on LogP. A lower value on these statistics for the molecules generated through two stages signals that they are much more similar to the test set on these properties. Structural statistics generally did not change a lot throughout the 2-stage and 3-stage training. The performance on these metrics of the later stages models may be bottle-necked by the first-stage graph decoder. We also repeat the second-stage VAE to perform the third-stage. However, there is no consistent improvement from the second stage across the board as seen in Table \ref{tab:chembl property table} and Table \ref{tab:polymer property table}. This is also in line with our synthetic experiment demonstrated in Figure \ref{fig:synthetic_sphere_mol}, where the second and the third stage of VAE training made no substantial improvement in manifold recovery compared to the improvements from the first stage to the second stage. Overall, the second stage VAE to the HGNN model outperforms both stages of RNN VAE and the latentGAN on majority of the evaluation metrics of both datasets.



The second stage to the RNN model does not provide significantly improvement on either datasets. The RNN VAE performs particularly poorly on the the polymer dataset as only half of the molecules the model generates are valid. This may be because that the RNN architecture is sensitive to the amount of data used for training. The polymer dataset is smaller than the ChEMBL dataset while polymers generally contain more atoms. This could negative impact RNN's performance. The second stage VAE slightly improves upon the first stage on the polymer datasets. On ChEMBL dataset, the second stage VAE does not have consistent improvement on any metrics. Our hypothesis for the poor performance of the 2-stage VAE with this RNN model as the first stage model is that the variance of the first stage decoder did not fulfill the condition of approaching 0 upon convergence. We will investigate the underlying reasoning behind this behavior for our future work.

\vspace{-8pt}
\section{Discussion}
\vspace{-8pt}
Manifold recovery is a challenge for VAE methods that train on data that lie on a low-dimensional manifold embedded in a higher-dimensional ambient space. In this paper, we presented a 2-stage VAE method that improves manifold recovery as demonstrated in a synthetic experiment. In experiments with molecular data such as ChEMBL and polymers, the method significantly improve the property statistics of a pre-existing VAE method and brings the generated molecules closer to the training dataset in property distributions. The nature of our approach makes it applicable to a wide range of other VAE based molecule generation methods. In future work we want to extend the method further to successfully adapt to more types of VAE architectures.

\newpage
\bibliographystyle{plainnat}
\bibliography{reference}
\newpage
\appendix
\section{Information on the Models Used for Experiments}

We chose one graph neural network based VAE and one RNN based VAE for diversity. We chose latentGAN as a non-VAE method for baseline.

\textbf{Hierarchical GNN}\citep{jin2019hierarchical}: The proposed method first extracts chemically valid motifs, or substructures, from the molecular graph such that the union of these motifs covers the entire molecule. The model consists of a fine-to-coarse encoder that encodes from atoms to motifs and a coarse-to-fine decoder that selects motifs to create the molecule while deciding the attachment point between the motif and the emerging molecule. We used the configuration from the orginal model. The latent size of the VAE is 32 and we used 0.1 as the KL coefficient.


\textbf{Vanilla RNN}\citep{polykovskiy2020molecular}: The inputs to the model are SMILES strings and the vocabulary consists of the low-level symbols in the SMILES strings. The encoder is a 1-layer GRU and the decoder is a 3-layer GRU. The latent size of the VAE is 128. We used most of the original configuration except the KL coefficient. 

\textbf{Latent GAN} \citep{prykhodko2019novo} is also a 2-stage method. The first stage is heteroencoder that takes SMILES strings as input while the second stage is a Wasserstein GAN with gradient penalty (WGAN-GP) that trains on the latents of the first stage VAE. The heteroencoder consists of an encoder and a decoder like an autoencoder and is trained with categorical cross-entropy loss. Afterwards, the GAN is trained to generate latent vectors for the decoder from the heteroencoder. We used the original parameters for training.

\section{Benchmark Metrics} \label{benchmark metric}
Property Statistics includes LogP (The Octanol-Water Partition Coefficient), SA (Synthetic Accessibility Score), QED (Quantitative Estimation of Drug-Likeness) and MW (Molecular Weight). These metrics determines the practicality of the generated molecules, for example, LogP measures the solubility of the molecules in water or an organic solvent \citep{wildman1999prediction}, SA estimates how easily the molecules can be synthesized based on molecule structures \citep{ertl2009estimation}, QED estimates how likely it can be a viable candidate of drugs \citep{bickerton2012quantifying}. The values listed in the table for each metric is the Wasserstein distance between the distribution of the property statistics in the test set and the generate molecule set. 

Structural statistics includes SNN (Similarity to Nearest Neighbor), Frag (Fragment Similarity), and Scaf (Scaffold Similarity). These statistics calculate two molecular datasets' structural similarity based on their extended-connectivity fingerprints \citep{rogers2010extended}, BRICS fragments \citep{degen2008art} and Bemis–Murcko scaffolds \citep{bemis1996properties}. 

The sample quality metrics are a lot more intuitive. Valid calculates the percentage of valid molecule outputs. Unique calculates the percentage of unique molecules in the first $k$ molecules where $k=1000$ for the ChEMBL dataset and $k=500$ for the polymer dataset. Novelty calculates the percentage of molecules generated that are not present in the training set. FCD is the Fr\'echet ChemNet Distance \citep{preuer2018frechet}.
\end{document}